\documentclass{article}

\usepackage{microtype}
\usepackage{graphicx}
\usepackage{subfigure}
\usepackage{booktabs} 

\usepackage{hyperref}

\usepackage[accepted]{mlsys2021}

\mlsystitlerunning{Semi-Supervised On-Device Neural Network Adaptation for Remote and Portable Laser-Induced Breakdown Spectroscopy}

\begin{document}

\twocolumn[
\mlsystitle{Semi-Supervised On-device Neural Network Adaptation for \\ Remote and Portable Laser-Induced Breakdown Spectroscopy}




\begin{mlsysauthorlist}
\mlsysauthor{Kshitij Bhardwaj}{llnl}
\mlsysauthor{Maya Gokhale}{llnl}
\end{mlsysauthorlist}

\mlsysaffiliation{llnl}{Lawrence Livermore National Laboratory, Livermore, CA}

\mlsyscorrespondingauthor{Kshitij Bhardwaj}{bhardwaj2@llnl.gov}


\vskip 0.3in

\begin{abstract}
{\em Laser-induced breakdown spectroscopy (LIBS)} is a popular, fast elemental analysis technique used to determine the chemical composition of target samples, such as in industrial analysis of metals or in space exploration. Recently, there has been a rise in the use of machine learning (ML) techniques for LIBS data processing. However, ML for LIBS is challenging as: (i) the predictive models must be lightweight since they need to be deployed in highly resource-constrained and battery-operated portable LIBS systems; and (ii) since these systems can be remote, the models must be able to self-adapt to any {\em domain shift} in input distributions which could be due to the lack of different types of inputs in training data or dynamic environmental/sensor noise. This on-device retraining of model should not only be fast but also unsupervised due to the absence of new labeled data in remote LIBS systems. We introduce a lightweight multi-layer perceptron (MLP) model for LIBS that can be adapted on-device without requiring labels for new input data. It shows 89.3\% average accuracy during data streaming, and up to 2.1\% better accuracy compared to an MLP model that does not support adaptation. Finally, we also characterize the inference and retraining performance of our model on Google Pixel2 phone.\vspace{-0.2in}
\end{abstract}
]
\printAffiliationsAndNotice{}

\section{Introduction}
\label{sec:intro}

Laser-induced breakdown spectroscopy (LIBS) is a fast elemental analysis technique with a variety of applications. LIBS involves analyzing the light spectrum obtained from a plasma explosion generated by the interaction of laser and the target material sample. This spectrum is analyzed to determine the composition of the sample, for example, the type of mineral. LIBS is used for industrial analysis of metals, geological research, and most notably in space exploration~\cite{libs-review}. In particular, remote and portable LIBS is employed in the SuperCam instrument on board the Mars Perseverance rover~\cite{supercam}.

With the advent of new more compact and robust laser technologies, there is a rise of portable LIBS systems. Such systems can be both deployed in remote areas (other planets, mines, under water) as well in handheld devices~\cite{portable-libs}. These LIBS devices also require fast on-device data processing to quickly analyze the generated spectra in situ (and in real time) and classify them in terms of the types of elements~\cite{portable-libs2}. For example, ThermoFisher's battery-operated LIBS device performs alloy identification in about 10 secs~\cite{tf-libs}.

Recent years have seen a rise in the use of machine learning (ML) techniques for processing of LIBS data to determine elemental compositions~\cite{libsML}. ML has been shown to be effective in handling high-dimensional and complex LIBS spectra, eliminating interference and noise, and improving the classification accuracy compared to other common statistical methods~\cite{libsML}. A variety of ML models have been used: random forests, support vector machines, as well as neural networks~\cite{svm-libs,libsML,emslibs}. 

However, designing ML models for portable and remote LIBS devices is challenging. In particular, there are two main challenges: (i) as these devices tend to be battery-operated, the model should be lightweight so it consumes less memory with reduced power consumption, as well as lead to fast real-time predictions while still able to achieve high accuracy; and (ii) the model must be able to efficiently handle {\em domain shift} that can occur during device operation and should continue to maintain prediction accuracy. This shift could be caused due to encountering spectral data that is different in distribution compared to the one used for offline training, as well as due to dynamic environmental or instrument sensor noise. Adapting an ML model to handle domain shift in an online setting is a difficult task especially when the labels for the shifted inputs are not available~\cite{nn-adapt, nn-adapt2}, which is usually the case for remote LIBS operations.
To address the above challenges, we make the following contributions.

We first introduce a new lightweight multi-layer perceptron (MLP) model, called {\em MLP-LIBS}, that achieves an average accuracy of 88.2\% on a well-known LIBS dataset with 12 mineral classes. The MLP-LIBS model consists of only two hidden layers with 64 neurons each. MLP-LIBS's accuracy is on par with the other ML models that have been introduced in a recent work for this dataset~\cite{emslibs}, which however do not handle domain shift.

We also extend MLP-LIBS to handle domain shift with very low overheads using an unsupervised technique based on standard backpropagation~\cite{nn-adapt}. The new MLP-LIBS-ADAPT model adds only two extra layers compared to MLP-LIBS. This model can be efficiently retrained on-device after deployment in a semi-supervised manner without the labels for the new domain-shifted inputs.

For a data streaming case study, MLP-LIBS-ADAPT shows up to 2.1\% accuracy improvement over MLP-LIBS that does not support adaptation to domain shift. We also measured the performance of MLP-LIBS-ADAPT on Google Pixel2 phone (Qualcomm Kryo 280 processor), where the average inference time for the model is 0.097 secs and average retraining time during adaptation is 599 secs.
Based on these results, we also propose a heterogeneous LIBS accelerator architecture to efficiently run MLP-LIBS-ADAPT.

To the best of our knowledge, we are not aware of any prior work on on-device neural network adaptation for LIBS and its performance characterization on an embedded processor. 

\vspace{-0.35in}
\section{LIBS background}
\vspace{-0.05in}
\label{sec:back}

This section briefly describes a portable LIBS system. It also provides details on the target LIBS dataset and examples of domain shift that can be found in the LIBS data. 

\begin{figure}
  \includegraphics[width=\columnwidth]{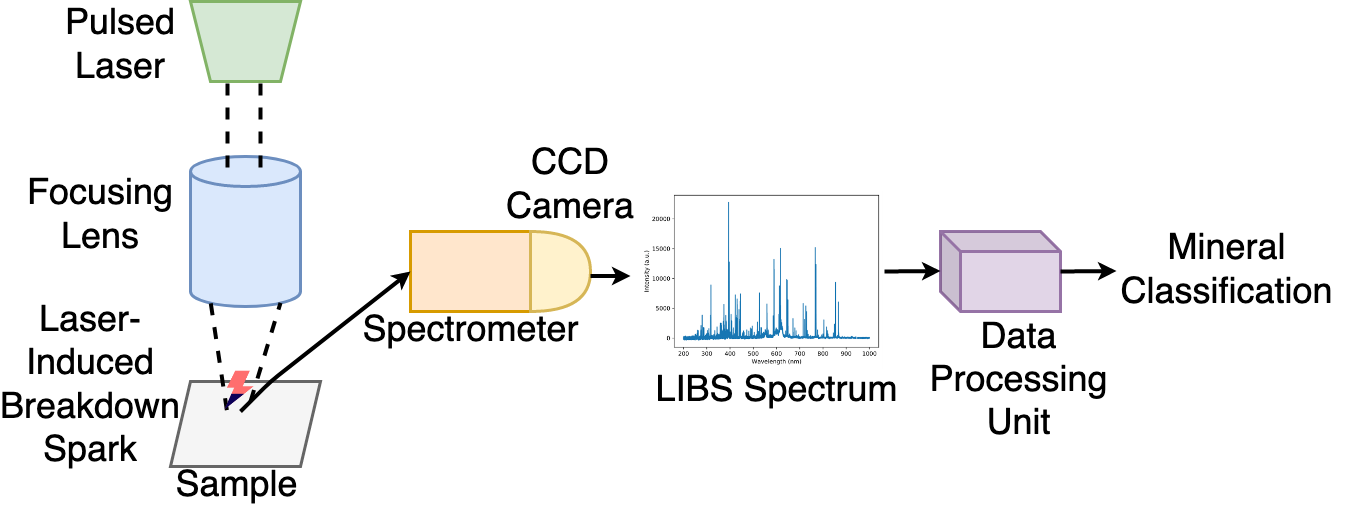}
  \vspace{-0.3in}
  \caption{Portable LIBS apparatus}
  \vspace{-0.1in}
  \label{fig:libs}
\end{figure}

\begin{figure}
\begin{center}
  \includegraphics[width=0.7\columnwidth]{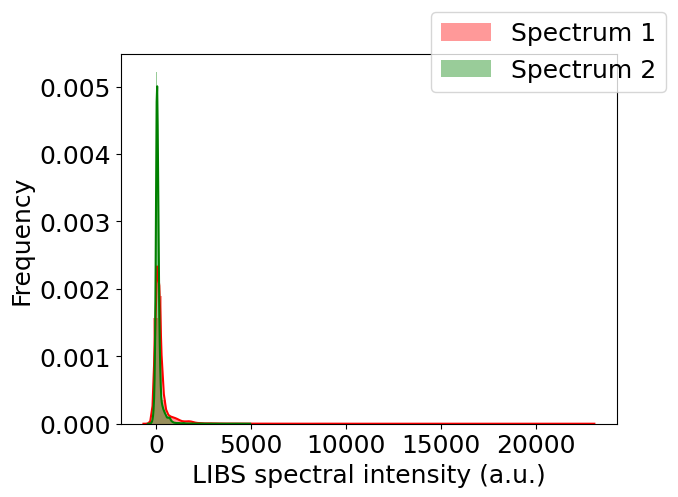}
\end{center}
  \vspace{-0.2in}
  \caption{Different probability distributions for same mineral class}
  \vspace{-0.25in}
  \label{fig:libs_dist}
\end{figure}

Figure~\ref{fig:libs} shows an example of a portable LIBS system~\cite{portable-libs}. It consists of a miniaturized pulsed laser, which concentrates a low-energy laser light on a sample using a focusing lens.
The high-temperature plasma is generated on the surface of the sample. After the plasma cooling process, light with different frequencies is radiated from the sample, which is collected through an optical fiber and sent through a spectrometer and a CCD camera. The final generated spectrum is used by an on-device data processing unit for qualitative and quantitative analysis based on the spectrum's wavelength and intensity.

In this paper, we use a well-known LIBS dataset~\cite{libs-dataset} for training and testing our MLP models. This dataset consists of 138 soil mineral samples, each belonging to one of 12 classes. Each sample consists of 500 spectra, where a spectrum has 40002 wavelength points. The data is based on an experiment conducted in a state-of-the-art LIBS interaction chamber. The samples were mapped with a $100 \mu m$ step size (distance between shots) at a 20 Hz ablation repetition rate with a pulse energy of 15 mJ. 

Figure~\ref{fig:libs_dist} shows examples of two spectra that belong to the same class of mineral in the above dataset but have considerably different probability distributions\footnote{also shown by a p-value of 0.0 in Kolmogorov-Smirnov test; $p-value < 0.01$ shows different distributions~\cite{ks}}. Such a difference in distributions can arise as the laser shots are targeted at different positions on the same sample material, which is usually the case in LIBS systems. If an ML model is trained with only samples similar to spectrum 1, it may incorrectly classify spectrum 2 during online operation. Similarly, environmental factors such as wind and temperature, and sensor noise can also introduce shifts in input distributions. Therefore, an ML model must be able to handle such domain shifts and adapt itself. Furthermore, such an adaptation needs to occur without any new labeled data due to the remote operation of the instrument.

\vspace{-0.2in}
\section{MLP model for LIBS}
\vspace{-0.05in}
\label{sec:model}

This section first describes the data preprocessing steps that were undertaken, followed by the proposed MLP model for the target LIBS dataset~\cite{libs-dataset}.


{\bf Data preprocessing.} As described above, each spectrum of the LIBS dataset consists of 40002 wavelength points and their corresponding intensity values. Since the intensities can vary significantly in magnitude (from negative to positive values) for various wavelength points of a spectrum, min-max normalization is performed to scale the values to [0,1] range. Furthermore, dimensionality reduction is also performed for each spectrum to reduce from 40002 dimensions to 100 features. In particular, uniform manifold approximation and projection (UMAP) is used as it has been shown to be fast, highly effective, and unsupervised technique for dimensionality reduction~\cite{umap}.

\begin{figure}
  \includegraphics[width=0.9\columnwidth]{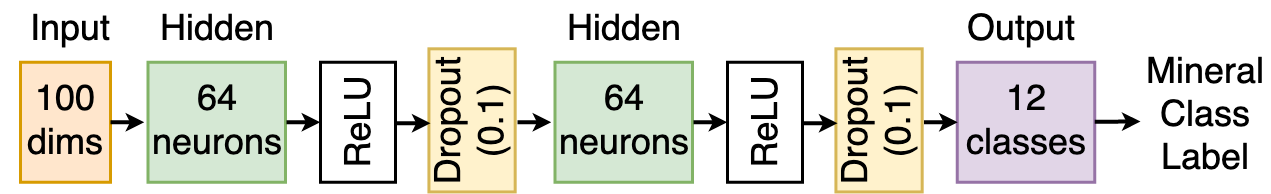}
  \vspace{-0.1in}
  \caption{MLP-LIBS model}
  \vspace{-0.2in}
  \label{fig:libs_mlp}
\end{figure}

{\bf MLP-LIBS model.} Figure~\ref{fig:libs_mlp} shows our MLP model for the LIBS data. The model's input layer takes a LIBS spectrum reduced to 100 dimensions. It then has two hidden layers, each with 64 neurons followed by ReLU activation function and a dropout layer to avoid overfitting. The final output layer selects one of the 12 mineral classes. Increasing the number of hidden layers (beyond 2) and number of neurons per layer (beyond 64) did not lead to improvement in model accuracy. Moreover, using a lightweight model enables memory and power savings for highly-constrained and battery-operated portable LIBS systems.

\vspace{-0.2in}
\section{Semi-supervised MLP adaptation}
\vspace{-0.05in}
\label{sec:adapt}

The lightweight architecture of the proposed MLP-LIBS-ADAPT and how it can be adapted on-device in a semi-supervised manner is described next.

\begin{figure}
  \includegraphics[width=0.9\columnwidth]{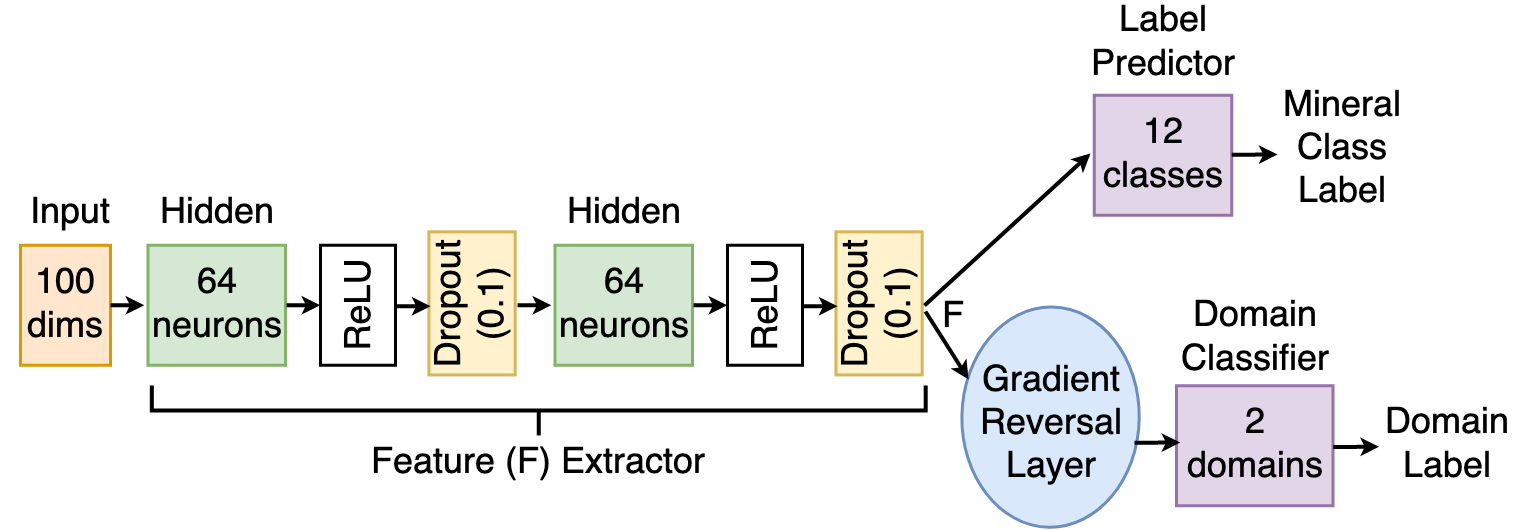}
  \vspace{-0.18in}
  \caption{MLP-LIBS-ADAPT model}
  \vspace{-0.25in}
  \label{fig:libs_adapt}
\end{figure}

{\bf Model architecture.} Figure~\ref{fig:libs_adapt} shows the proposed MLP-LIBS-ADAPT architecture. This model extends MLP-LIBS using an unsupervised domain adaptation technique based on standard backpropagation~\cite{nn-adapt}. There are only two minor additions compared to MLP-LIBS: (i) a new gradient reversal layer; and (ii) a domain classifier that classifies the input spectrum as: from the {\em source} domain that corresponds to labeled data that the model was trained on, or {\em target} domain that represents new unlabeled input data that may be domain shifted.

{\bf On-device model adaptation.} Before deployment, MLP-LIBS-ADAPT model is first trained offline on a set of labeled LIBS data (i.e the source domain) using the standard backpropagation algorithm based on stochastic gradient descent. After the model is deployed on a device, it can be adapted (or retrained) at user-defined instances, such as when the instrument is dormant. For each retraining, two sets of data are used: (i) {\em labeled long term memory (LLTM)} which is the spectra samples with mineral labels that the model was trained with initially (source domain); and (ii) {\em unlabeled short term memory (USTM)} that is the recently seen, possibly domain shifted, spectra samples without any mineral labels (target domain). Both the data sets are assigned domain pseudo-labels, LLTM as 0 and USTM as 1, followed by training using backpropagation. 

During on-device training, the model learns domain-invariant features between the source and target domains which enable it to make classification decisions without being hindered by the shift between the two domains.
This learning is achieved by the gradient reversal layer that multiplies the backpropagation gradient with a negative constant, which is varied between -1 to 0 as the training epochs advance. Therefore, the model learns those features that minimize the label prediction loss for LLTM samples (i.e. the features are discriminative) but maximizes the domain prediction loss for USTM samples (i.e. extracting those features that are domain-invariant). However, when performing the forward pass, the gradient reversal layer leaves the input unchanged. More details are in~\cite{nn-adapt}.

\vspace{-0.3in}
\section{Results}
\vspace{-0.05in}
\label{sec:results}

The effectiveness of MLP-LIBS-ADAPT in terms of model adaptation over MLP-LIBS is evaluated. This section also presents performance measurements of running MLP-LIBS-ADAPT on Google Pixel2 phone.

\vspace{-0.2in}
\subsection{Model adaptation evaluation}
\vspace{-0.05in}
\label{subsec:online_results}

{\bf Setup.} The LIBS dataset~\cite{libs-dataset} as described in Section~\ref{sec:back} is used in these studies for training and testing MLP-LIBS and MLP-LIBS-ADAPT. Both these models are initially trained and validated offline. The amount of the initial training data is varied with three cases considered: 20000, 18000, and 4000 spectra. For online adaptation of MLP-LIBS-ADAPT, there are two parameters: amount of LLTM and USTM data. The number of labeled spectra used as LLTM also varies and it could be the same as the offline training data or some fraction of this data. Further, two cases of USTM are used: recently seen 50 or 100 unlabeled spectra. Finally, a separate set of 25000 test spectra are used to simulate data streaming. During this streaming, prequential evaluation (interleaved test-then-train)~\cite{preq} is used for MLP-LIBS-ADAPT, where its test accuracy is reported for every 2500 spectra or data chunks (i.e. 10 times), followed by retraining it using the most recent USTM and pre-stored LLTM. MLP-LIBS, on the other hand, is not retrained during the test stream. All model training and testing are performed using PyTorch~\cite{pt}. 

\begin{figure}
  \includegraphics[width=\columnwidth,trim=4 4 4 4,clip]{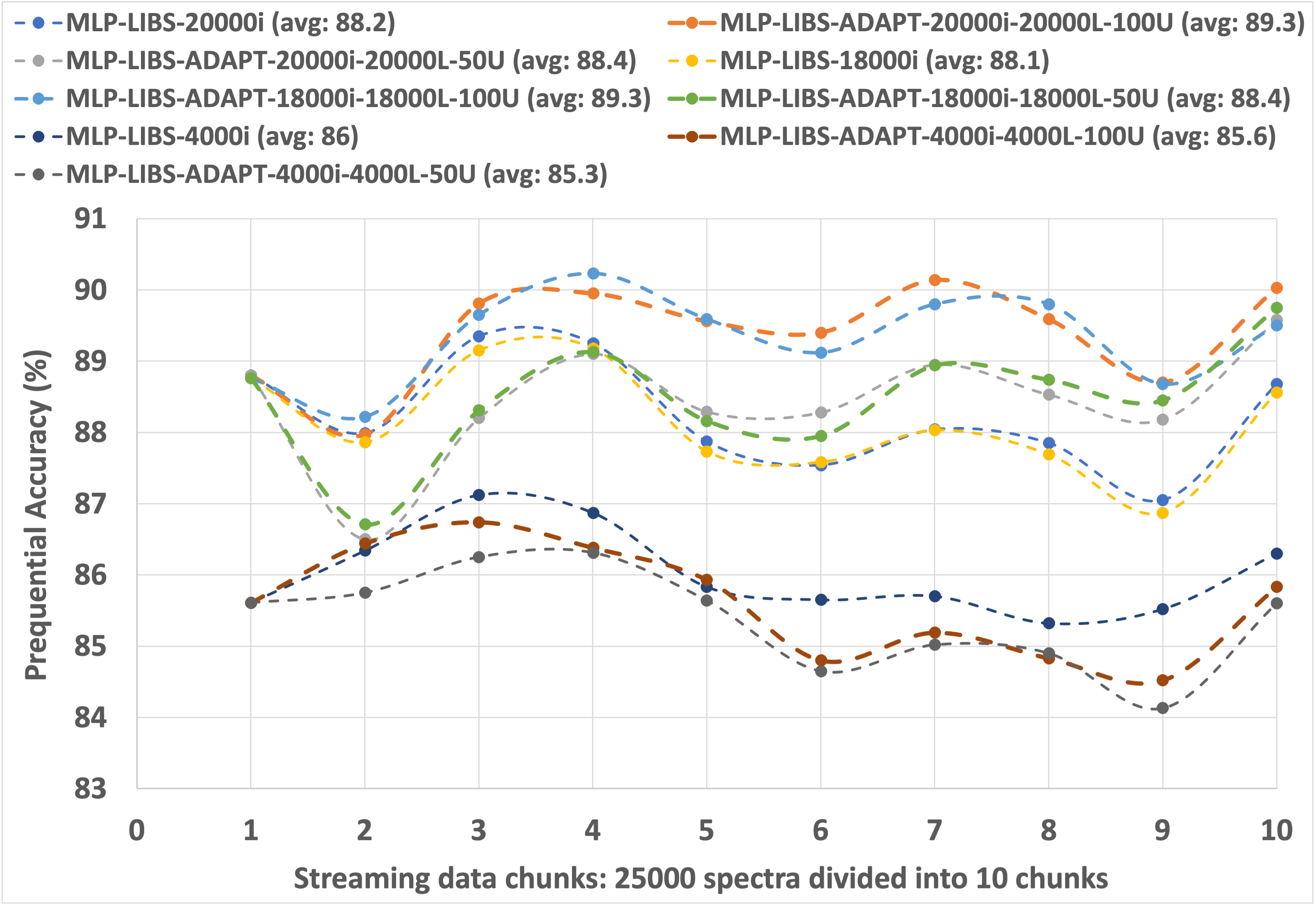}
  \vspace{-0.3in}
  \caption{MLP-LIBS vs. MLP-LIBS-ADAPT}
  \vspace{-0.15in}
  \label{fig:libs_res1}
\end{figure}

{\bf Results.} Figure~\ref{fig:libs_res1} demonstrates the effectiveness of MLP-LIBS-ADAPT over MLP-LIBS during data streaming, while varying three parameters: initial offline training data, LLTM, and USTM (shown as i, L, U, respectively).
Both i and L are fixed to be the same in this study. Each of the 9 combinations was run multiple times and the average accuracy numbers are reported across these runs for the 10 different data points. The legend also presents the accuracy, averaged over the 10 data chunks, for each of the 9 cases. 

As expected, more amount of initial training data leads to better accuracy (20000/18000 vs. 4000). However, 20000 and 18000 training spectra lead to similar accuracy for this test set. MLP-LIBS-20000i shows an average accuracy of 88.2\% that is similar to that of the other ML models developed for this dataset~\cite{emslibs}, which however do not support adaptation. Our MLP-LIBS-ADAPT using 20000L-100U shows 1.1\% average improvement over MLP-LIBS, and 1.2\% average improvement when using 18000L-100U. The best improvement is shown for data chunk 7 where MLP-LIBS-ADAPT (using 20000L-100U) outperforms MLP-LIBS by 2.1\%. 
However, reducing the USTM of MLP-LIBS-ADAPT to 50 leads to lower accuracy compared to 100U by an average of 0.9\% for both 20000 and 18000 cases. Interestingly, when using 4000 spectra for training, MLP-LIBS-ADAPT performs worse than MLP-LIBS. Therefore, if the initial training data (and LLTM) is too small, online adaptation can lead to the opposite effect of worsening the model performance.

\begin{figure}
  \includegraphics[width=\columnwidth,trim=4 4 4 4,clip]{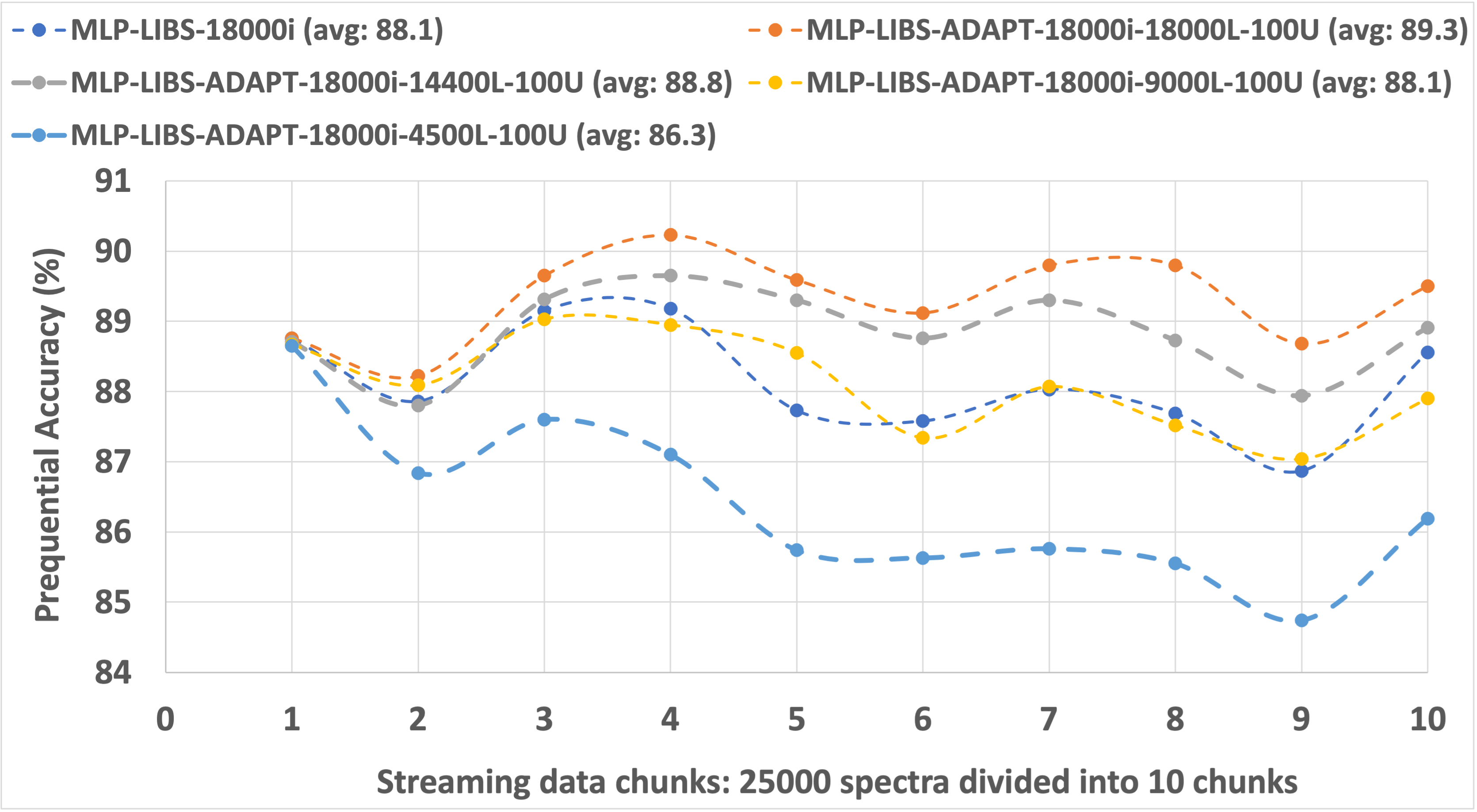}
  \vspace{-0.3in}
  \caption{Effect of using fractions of initial training data as LLTM}
  \vspace{-0.25in}
  \label{fig:libs_res2}
\end{figure}

While the above results used the same amount of LLTM as the initial offline training data, Figure~\ref{fig:libs_res2} shows the effect of using different fractions of initial data as LLTM. In this case and without loss of generality, the initial training data is fixed to be 18000, whereas LLTM is varied to be: 14400, 9000, and 4500. USTM is also fixed to be 100. As evident, the accuracy of MLP-LIBS-ADAPT suffers as LLTM amount is reduced. These results show that not only LLTM should be significant, it should also be ideally equal to the initial training data for best on-device model adaptation.

\vspace{-0.2in}
\subsection{Model performance on Google Pixel2}
\vspace{-0.07in}
\label{subsec:pixel_results}

We ran multiple configurations of MLP-LIBS-ADAPT on Pixel2, which uses Qualcomm's Kryo 280 processor. Considering the trade-offs between accuracy, memory requirement, and MLP processing time, the optimal configuration selected is 18000i-18000L-100U as it achieves an average accuracy of 89.3\% during model adaptation with average retraining time (across the 10 streaming data chunks) as 599 secs, and average inference time for each of the data chunks as 0.097 secs. While 20000i-20000L-100U also shows the same average accuracy, it requires a larger LLTM than the 18000 case as well as shows a longer training time on Pixel2: 682.2 secs. Finally, while 4000i-4000L-100U requires a very small LLTM and has a training time of only 126.5 secs, it suffers from a significant accuracy loss (85.6\%).

\vspace{-0.2in}
\section{Accelerator Design for LIBS}
\vspace{-0.05in}
\label{sec:hw_arch}

\begin{figure}
  \includegraphics[width=0.9\columnwidth]{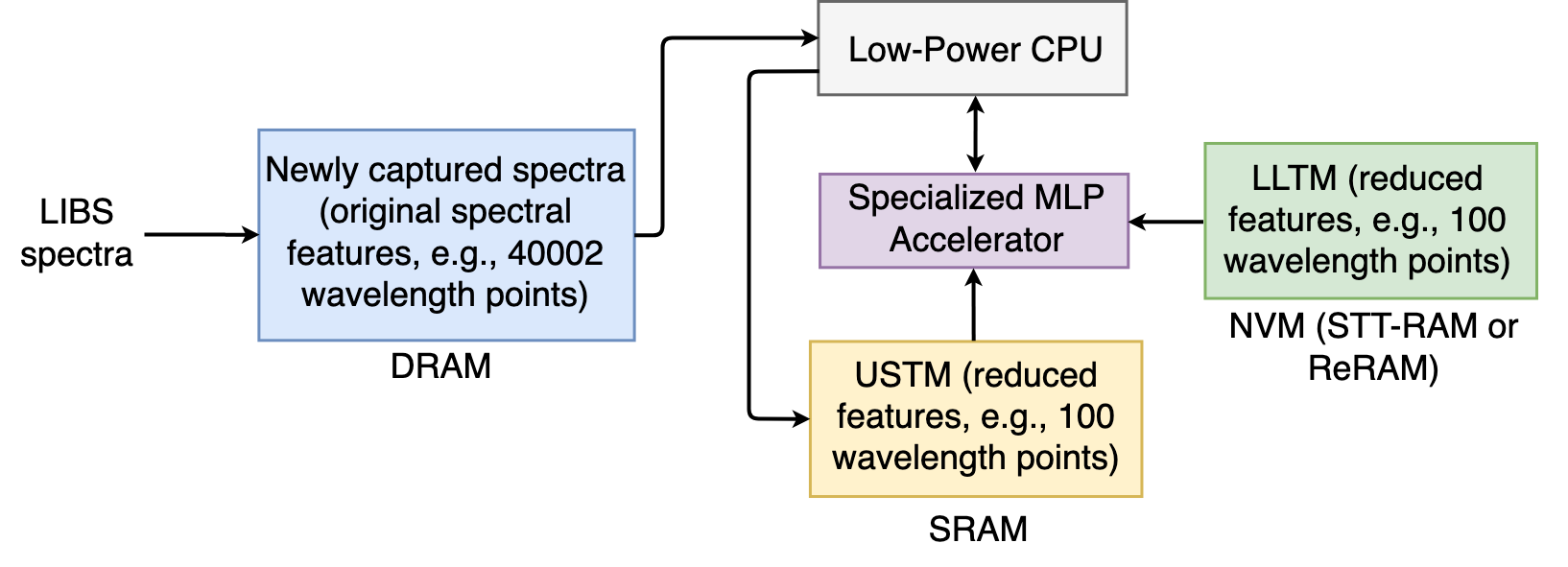}
  \vspace{-0.22in}
  \caption{Proposed accelerator architecture for LIBS}
  \vspace{-0.3in}
  \label{fig:libs_acc}
\end{figure}

The above results demonstrate the need for a heterogeneous hardware architecture for portable LIBS applications. Figure~\ref{fig:libs_acc} shows our proposed accelerator architecture, which will benefit from heterogeneity, both in terms of different memory technologies as well as a specialized MLP training accelerator. The implementation and evaluation of this architecture is left as future work.

The newly captured LIBS spectra during instrument operation, with original high-dimensional wavelength points, will be first stored in a DRAM. Further, semi-supervised online adaptation of our MLP model requires two kinds of memory: LLTM and USTM. Ideally, LLTM should contain all the samples that the model was trained with during initial offline training. This read-only memory can be an emerging non-volatile memory (NVM), such as STT-RAM or ReRAM, due to its advantages of high-density storage and fast read accesses. USTM, on the other hand, can simply be a scratchpad SRAM storing a window of recently captured spectra after their dimensions have been reduced. This USTM will be used for both inference and model adaptation.

In terms of computing, a low-power CPU will be used for initiating data movement, performing data preprocessing tasks (normalization and dimensionality reduction), and invoking a specialized MLP accelerator. The accelerator is assumed to be able to support high-throughput, low-latency inference as well as on-device training of MLPs, similar to some of the recent advances~\cite{nn-acc}. 

\vspace{-0.2in}
\section{Conclusion and Future Work}
\vspace{-0.05in}
\label{sec:conclusion}

This paper introduces lightweight MLP-LIBS-ADAPT for portable and remote LIBS systems, which can also adapt to any domain shift in a semi-supervised manner. MLP-LIBS-ADAPT is shown to achieve $\sim90\%$ accuracy during data streaming, and up to 2.1\% better accuracy than a model which does not adapt. Our experiments show that for effective model adaptation, labeled long term memory should be ideally equal to the initial offline training data, and the recently seen unlabeled short term memory should also be sufficient. We also characterized inference and retraining performance of MLP-LIBS-ADAPT on Google Pixel2. As a future work, we plan to prototype a custom heterogeneous system-on-chip to efficiently run MLP-LIBS-ADAPT.
\section{Auspices and Disclaimers}
\label{sec:ad}

Prepared by LLNL under Contract DE-AC52-07NA27344.
This document was prepared as an account of work sponsored by an agency of the United States government. Neither the United States government nor Lawrence Livermore National Security, LLC, nor any of their employees makes any warranty, expressed or implied, or assumes any legal liability or responsibility for the accuracy, completeness, or usefulness of any information, apparatus, product, or process disclosed, or represents that its use would not infringe privately owned rights. Reference herein to any specific commercial product, process, or service by trade name, trademark, manufacturer, or otherwise does not necessarily constitute or imply its endorsement, recommendation, or favoring by the United States government or Lawrence Livermore National Security, LLC. The views and opinions of authors expressed herein do not necessarily state or reflect those of the United States government or Lawrence Livermore National Security, LLC, and shall not be used for advertising or product endorsement purposes.

\bibliography{main}
\bibliographystyle{mlsys2021}

\end{document}